\documentclass{bmvc2k}
\usepackage{times}
\usepackage{epsfig}
\usepackage{graphicx}
\usepackage{amsmath}
\usepackage{amssymb}
\usepackage{booktabs}
\usepackage{mathtools}
\usepackage{layouts}

\usepackage{multirow}
\usepackage{enumitem}
\usepackage{color}

\newcommand{\keypoint}[1]{\vspace{0.1cm}\noindent\textbf{#1}\quad}
\newcommand{\cut}[1]{}
\usepackage{algorithm}
\usepackage{algpseudocode}

\usepackage[nopar]{lipsum}
\usepackage{arydshln}
\DeclareMathAlphabet\mathbfcal{OMS}{cmsy}{b}{n}

\def\black#1{\textcolor[rgb]{0,0,0}{#1}}

\usepackage{array}
\usepackage{colortbl}
\usepackage{hhline}
\usepackage{soul}
\usepackage{pgfplots}
\usepackage{wrapfig}
\usepackage{pgf}
\pgfplotsset{width=10cm,compat=1.9}
\usepackage[colorlinks]{hyperref}

\title{Active Learning for Fine-Grained Sketch-Based Image Retrieval}

\addauthor{Himanshu Thakur}{hthakur@andrew.cmu.edu}{$\ast$ 1}
\addauthor{Soumitri Chattopadhyay}{soumitri@cs.unc.edu}{$\ast$ 2}

\addinstitution{Carnegie Mellon University} 
\addinstitution{UNC Chapel Hill} 

\runninghead{Thakur, Himanshu and Chattopadhyay, Soumitri}{AL4FGSBIR}

\begin{document}

\maketitle






\begin{abstract}

\black{The ability to retrieve a photo by mere free-hand sketching highlights the immense potential of Fine-grained sketch-based image retrieval (FG-SBIR). However, its rapid practical adoption, as well as scalability, is limited by the expense of acquiring faithful sketches for easily available photo counterparts. A solution to this problem is Active Learning, which could minimise the need for labeled sketches while maximising performance. Despite extensive studies in the field, there exists no work that utilises it for reducing sketching effort in FG-SBIR tasks. To this end, we propose a novel active learning sampling technique that drastically minimises the need for drawing photo sketches. Our proposed approach tackles the trade-off between uncertainty and diversity by utilising the relationship between the existing photo-sketch pair to a photo that does not have its sketch and augmenting this relation with its intermediate representations. Since our approach relies only on the underlying data distribution, it is agnostic of the modelling approach and hence is applicable to other cross-modal instance-level retrieval tasks as well. With experimentation over two publicly available fine-grained SBIR datasets ChairV2 and ShoeV2, we validate our approach and reveal its superiority over adapted baselines.
}
\end{abstract}

\def\thefootnote{$\ast$}\footnotetext{Both authors contributed equally to the paper.}\def\thefootnote{\arabic{footnote}}

\section{Introduction}



The success of computer vision applications can be largely attributed to deep learning architectures \cite{huang2017densely, szegedy2015going}, which, in turn, have yielded favourable results due to their access to large-scale labelled databases \cite{deng2009imagenet, lin2014microsoft} for training. Being in the age of Big Data, enormous volumes of data is easily available; however, proper annotation of the same is a painstakingly cumbersome as well as an expensive process, often requiring specialized qualifications if the task at hand demands for domain expertise, such as handling medical images \cite{wu2021strongly}. To alleviate this bottleneck, researchers have proposed various annotation-efficient methods \cite{beluch2018power, li2019label, morgado2021audio, zhang2020state} to standard computer vision tasks like classification and segmentation. A commonly used technique is active learning \cite{sener2017active, sinha2019variational, budd2021survey}, which seeks to find the most ``useful" unlabelled data samples to be annotated for learning, so as to reduce annotation cost as well as increase overall generalisability on the supervised learning task to be performed. Apart from conventional visual tasks, active learning has been applied to other domains such as video captioning \cite{chan2020active}, hand pose estimation \cite{caramalau2021active} and single-image super-resolution \cite{wang2016single}.




\begin{wrapfigure}{r}{0.5\textwidth}
\vspace{0cm}
\centering
    \includegraphics[width=0.5\textwidth]{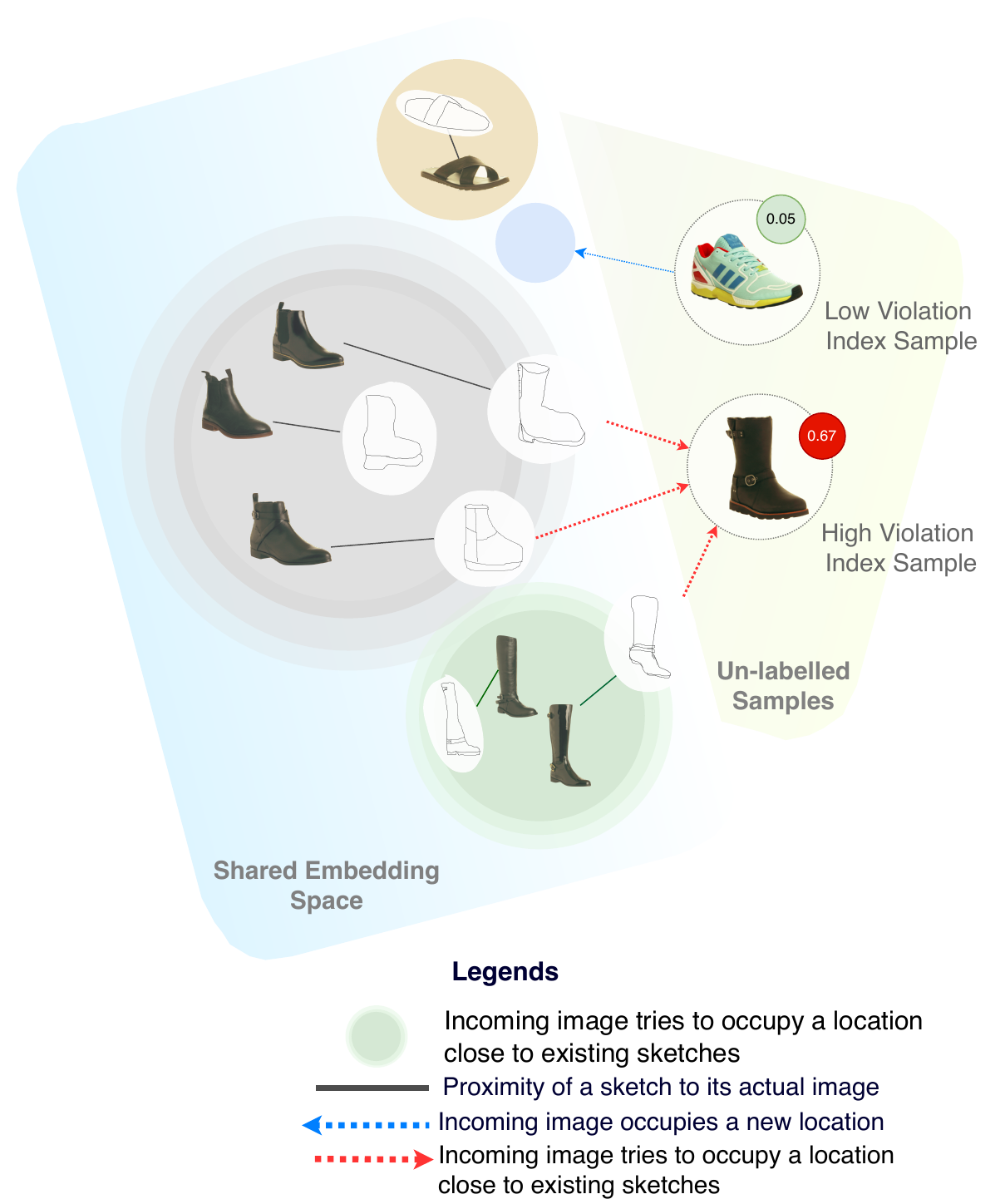}
\vspace{-0.6cm}    
\caption{Intuition behind our proposed AL framework. The \textit{violation index} quantifies the ``disturbance'' introduced into the learned embedding space by the incoming photo sample, due to having a greater similarity with a previously paired sketch present in the latent space. More details are provided in \autoref{sec:method}.}
    \label{fig:teaser}
\end{wrapfigure}

In this paper, we embrace a paradigm shift to tackle the aforementioned challenges in a domain which is fundamentally very much different from traditional vision tasks -- fine-grained sketch-based image retrieval (FG-SBIR) \cite{bhunia2020sketch, bhunia2021more, sain2021stylemeup}, a relatively newer direction of research from traditional category-level SBIR \cite{wang2015sketch, collomosse2017sketching}. As the name suggests, FG-SBIR aims at exploiting the finer sketch representations for cross-modal instance-level retrieval, achieved by learning an embedding space where sketch-photo pairs lie close to each other. The most common approach in several works \cite{yu2016sketch, pang2019generalising, bhunia2020sketch} has been to train a supervised triplet loss-based model \cite{hoffer2015deep} that learns feature similarities between an image and its corresponding sketch and hence requires a large number of sketch-photo pairs. However, drawing full sketches is both time-consuming and difficult, since it requires artistic expertise and amateurish sketching can only lead to learning degradation. Thus, there is a need to develop a robust, annotation-efficient pipeline for FG-SBIR. Very few recent works have been proposed treading on this motivation, such as a generalisable zero-shot \cite{pang2019generalising} and semi-supervised learning \cite{bhunia2021more} FG-SBIR framework. While the latter involves training of two networks which makes it computationally expensive, the performance of \cite{pang2019generalising} is far from fully-supervised alternatives.


\black{Developing an Active Learning pipeline for FG-SBIR imposes some unique challenges. Considering a traditional FG-SBIR model which lacks a probability distribution of the samples in its output, there is an absence of a direct way to measure the uncertainty of such a model. Hence, it becomes difficult to select a photo for labelling without having an estimate of its uncertainty from the model.} \black{Moreover, off-the-shelf active learning methods \cite{sinha2019variational, kirsch2019batchbald} that were primarily proposed for classification tasks are not suitable for FG-SBIR, since for classification the learning mechanism \textit{draws firm discriminatory boundaries} among samples, whereas in a cross-modal instance-level retrieval setup the objective is to draw \textit{softer decision boundaries}, as the samples belong to the same category and only differ in minute fine-grained details.} \black{Additionally,  the FG-SBIR model holds the photo and sketch embeddings in a joint space and thus, selection based on only photo or sketch might yield previously unseen results, compared to selection from a single modality. Hence, developing a sampling technique for the FG-SBIR requires the handling of both modalities - photo and its sketch.} 


The effectiveness of an active learning pipeline depends highly upon the \textit{technique} of selecting samples from the unlabelled pool. As a result, a good technique for sampling photos would lead to the maximum increase in the model's performance. To this end, we propose a novel sampling strategy for active learning that utilises the evolving relations between photos and their sketches to approximate the influence of a new photo from the unlabelled pool, on the model's existing knowledge. Our sampling technique is informed by the embedding space learnt using the labelled pool of photo-sketch pairs and two major aspects of an unlabelled photo -- \textit{its predicted representation} and \textit{its approximate potential of influence} in the existing embedding space. While the former is the basis of our sampling technique, the latter helps in tackling the classic trade-off between uncertainty and diversity-based sampling techniques. Specifically, to model an unlabelled photo's potential of influencing the existing embedding space, we formulate a quantity, \textit{violation index}. The violation index of a photo acts as a proxy for measuring the amount of confusion it could create in the existing sketch-photo pair embedding space (refer to \autoref{fig:teaser} for an intuitive understanding). \black{Further, for diversity sampling, we choose $k$-MEANS++ due to its ability to converge faster, as highlighted in \cite{ash2019deep}. The overall advantage is two-fold: since we adopt a relative method of approximating the influence using the task network itself, it does not need an auxiliary learner or source of knowledge. Moreover, our technique relies on the underlying distribution of the data itself and hence is agnostic of the model used. The number of samples to be chosen depends on the permissible budget of annotation; the chosen samples are then queried for their sketches to be drawn, which are then paired and put into the training examples. }

To sum up, the primary contributions of the presented study are as follows: (1) For the first time, we propose an \textit{active learning} pipeline for annotation-efficient fine-grained SBIR; (2) To this end, we formulate a novel sampling strategy that incorporates uncertainty as well as diversity to quantify the ``usefulness" of an unlabelled sample for querying its label (here, sketch); (3) With suitable experimentation and comparison with adopted baselines on two publicly available FG-SBIR datasets, as well as conducting ablations on the proposed framework, we demonstrate the usefulness of our approach.

\section{Related Works}


  
  
\keypoint{Fine-grained SBIR:} Although SBIR was originally proposed and studied as a category-level retrieval task \cite{wang2015sketch, bui2018deep, sain2022sketch3t}, recently there has been significant interest among researchers towards exploiting the \emph{fine-grained} information that sketches provide \cite{bhunia2020sketch, song2017deep, pang2019generalising, bhunia2021more, sain2023exploiting} for enhanced cross-domain matching, something other query mediums (e.g. text) fail to do. The first deep learning-based approach by Yu \textit{et al.} \cite{yu2016sketch} was further enhanced using cross-domain generative-discriminative learning \cite{pang2017cross} and attention mechanisms \cite{song2017deep}. More recent studies include zero-shot-like cross-category FG-SBIR \cite{pang2019generalising, sain2023clip}; cross-modal co-attention-based hierarchical model \cite{sain2020cross}; on-the-fly SBIR setup for early retrieval \cite{bhunia2020sketch}; style-agnostic meta-learning setup \cite{sain2021stylemeup} and a semi-supervised framework \cite{bhunia2021more} to tackle data scarcity in FG-SBIR \cite{sain2023exploiting}. However, none of these works seeks to address the need for a smart data labelling system for cross-modal instance-level retrieval problems like FG-SBIR so that we can achieve optimal performance using the minimum possible labelling budget.

\keypoint{Active Learning:} Active learning (AL) \cite{ren2020survey, budd2021survey} has been extensively studied for over two decades, the primary goal being to develop an effective strategy to reduce annotation effort by ``actively" selecting representative samples and improve learning. Apart from conventional image classification \cite{beluch2018power, sinha2019variational}, AL has been widely used for various computer vision tasks such as medical imaging \cite{budd2021survey}, image and video segmentation \cite{siddiqui2020viewal, xie2020deal, belharbi2021deep}, among others. Broadly speaking, AL approaches may be categorized as: (1) Uncertainty-based methods \cite{wang2014new, beluch2018power, yoo2019learning, choi2021vab}, which aim to construct a so-called acquisition function that quantifies the uncertainty of the model on unlabelled data points, based on which the points are sampled and queried for annotation; and (2) Representation-based methods \cite{sener2017active, sinha2019variational, ash2019deep, kim2021task}, which seek to learn a common embedding space for the labelled and unlabelled data items so as to sample the unlabelled data points that capture the most diverse regions of the embedding and thereby better represent the overall data distribution. Existing AL methods mostly deal with classification and thus cannot be directly adopted for FG-SBIR where paired photo and sketch need to be aligned in the vicinity in the joint embedding space. This cross-modal instance-wise matching brings an exclusively different set of challenges for employing AL in FG-SBIR. We intend to address and tackle these through this work, which, to the best of our knowledge, is the first to introduce AL to a cross-modal instance-level retrieval problem.     



\section{Background and Problem Formulation}\label{sec:background}

\keypoint{Baseline FG-SBIR:}Instead of complicated pre-training \cite{pang2020solving} or joint-training \cite{bhunia2021more}, we use a three-branch state-of-the-art siamese network \cite{pang2019generalising} as our baseline retrieval model, which is considered to be a strong baseline to date \cite{pang2019generalising, song2017deep}. Each branch starts from ImageNet pre-trained VGG-16 \cite{simonyan2015very}, sharing equal weights. Given an input image $I \in \mathbb{R}^{H \times W \times 3}$, we extract the convolutional feature-map $\mathcal{F}({I})$, which upon global average pooling followed by  $l_2$ normalisation
generates a $d$ dimensional feature embedding. This model has been trained with an anchor sketch (a), a positive (p) photo, and a negative (n) photo triplets $\{a,p,n\}$ using \textit{triplet-loss}. Triplet-loss aims at increasing the distance between anchor sketch and negative photo $\delta^{-}={||\mathcal F(a)-\mathcal F(n)||}_2$, while simultaneously decreasing the same between anchor sketch and positive photo $\delta^{+}={||\mathcal F(a)-\mathcal F(p)||}_2$. Therefore, the triplet-loss $\mathcal{L}$ with the margin hyperparameter $\mu>0$ can be written as:

\begin{equation}
\label{eq1}
    \mathcal{L}=max\{0, \delta^{+}-\delta^{-}+\mu\}
\end{equation}

During inference, given a gallery of $M$ photos $\{P_i\}_{i=1}^{M}$, we can compute a list of $d$ dimensional vectors as $G = \{\mathcal{F}(P_i)\}_{i=1}^{M}$. Now, given a query sketch $S$, and pair-wise distance metric, we obtain a top-q retrieved list as $Ret_q(F(S), G)$. If the paired (ground-truth) photo appears in the top-q list, we consider accuracy to be true for that sketch sample. 


\keypoint{Active Learning for FG-SBIR:}Following the principle of active learning shown in \autoref{eq:actL}, we aim to minimise the number of rounds $\mathcal{R}$ so that the fewest possible sketches are needed to be acquired. At the end of each of round $r$, the performance measures $\mathcal{P}$  are recorded to quantify and understand the effectiveness of our sampling technique $\mathcal{X}$.

\begin{equation}\label{eq:actL}
\min_{\mathcal{R}} \min_{\mathcal{L}} {[\mathcal{X}(\mathcal{{L}}|\mathcal{P}_0\subset \cdots \mathcal{P}_k \subset \mathcal{P}_U)]}_{r=1}^{\mathcal{R}}
\end{equation}

\section{Proposed Method}\label{sec:method}

\begin{figure*}[h]
    \centering
    \includegraphics[width=1\textwidth]{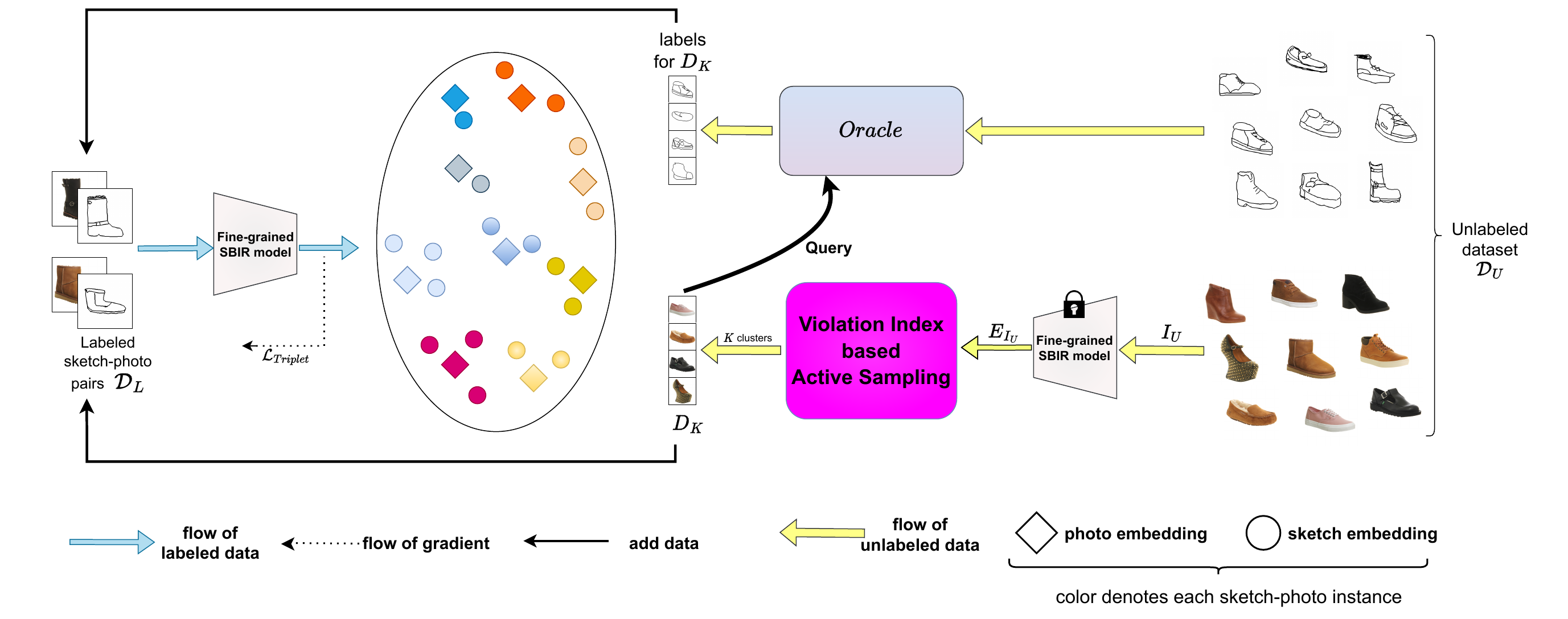}
    \vspace{-0.5cm}
    \caption{Overall workflow of the proposed active learning framework for fine-grained SBIR. The method starts with a subset of sketch-photo pairs for learning the FG-SBIR model, following which it is used to compute violation index of unlabeled photos with respect to the embedding space. Our sampling technique selects a suitable query set of photos which are passed to the Oracle (the ground truth sketch provider for queried photos) to obtain their sketch counterparts. These pairs are subsequently added to the labeled subset.}
    \label{fig:mainfig}
    \vspace{-0.2cm}
\end{figure*}

\vspace{-0.2cm}
\keypoint{Overview:}We aim to design an active learning framework specially suited for fine-grained SBIR so as to make it label-efficient. To keep our framework fairly simple, we have used the baseline triplet network (described in Section \ref{sec:background}) as the retrieval model. During the training process, we leverage active learning to select the most informative examples for querying its labels, the details of which are given in the subsequent sections. The AL mechanism is employed in training cycles; at the end of each cycle the sampling technique selects photos from the unlabelled data and queries for their paired sketches, which are provided by the Oracle (ground truth annotator) and added to the labelled training pool.


\subsection{Introducing Violation Index}

In this section, we introduce the core idea of our contribution -- the {\bf violation index}. Our approach is inspired by the learning process of cross-modal retrieval frameworks. When an FG-SBIR model learns, it tries to put an image and its sketch nearby in the embedding space, while pushing away other images. Formally, consider a model $\mathcal{M}$, the embedding of an image obtained from final layer of the model as ${E_{I}}$, and that of the corresponding sketch as ${E_{S}}$. Also, let ${E_{I'}}$ be the embedding of an image other than ${E_{I}}$, $N_{L}$ be the total size of the labelled dataset $\mathcal{D}_L$ and $N_{U}$ be the total size of the unlabeled dataset $\mathcal{D}_U$. To learn the similarities, following Section \ref{sec:background} let us consider a triplet loss function defined as follows: 

\begin{equation}
    {\mathcal {L}} = \operatorname {max} \left({\|E_{I} - E_{S}\|}^{2}-{\|E_{I} - E_{I'}\|}^{2}+\mu ,0\right)
\end{equation}

The objective of the learning process is to minimise $\mathcal{L}$ over all images in the dataset (Eq \ref{eq:2}). Hence, an ideal embedding space would comprise images lying closest to its actual sketch. However, a non-ideal embedding space is more realistic and introduces some new challenges.

\begin{equation}\label{eq:2}
 {minimize} \sum _{i=1}^{{}N_{L}}{\mathcal {L}}\left(E_{I}^{(i)},E_{S}^{(i)},E_{I'}^{(i)}\right)
\end{equation}

Now, when a new image $I'$ from the unlabelled set $\mathcal{D}_U$ is introduced during training, its corresponding sketch might not lie closest to it in the embedding space, a condition defined in \autoref{eq:3}. Hence, in the model’s view, the image now seems closer to one of the existing sketches rather than its own. This phenomenon leads to perturbation in the existing embedding space. We quantify the degree of disturbance an image from the unlabelled pool of images produces in the existing embedding space and call it its \emph{violation index}.

\begin{equation}\label{eq:3}
    {\|E_{I'} - E_{S}\|}^{2} \leq  {\|E_{I} - E_{S}\|}^{2}\
\end{equation}

Formally, we define the violation index ($VI$) of the photo embedding $E_{I'}$ as:

\begin{equation}\label{eq:4}
  Violation\,Index(E_{I'}) = \frac{1}{N_{L}} \times \sum _{i=1}^{{}N_{L}}\frac {\|E_{I}^{(i)} - E_{S}^{(i)}\|}{\|E_{I'} - E_{S}^{(i)}\|}
\end{equation}

Thus, the violation index is an improvement over a simpler distance-based sampling technique since it accounts for the fact that the inherent imperfections in sketches cause the cross-modal embedding space to be sensitive to new image-sketch pairs. Moreover, since the metric is built on relative distances instead, computing the relative
similarities between a new image and existing pairs help
surface novel samples that still do not violate existing sketches
or images.

\subsection{VI-based Active Sampling}

Given the violation index of an image shows its average relative distance from existing image-sketch pairs and hints towards how many of them it violates in the learned embedding space. Intuitively, images with a low violation index would be relatively unseen in the training data, whereas ones with a high violation index might closely resemble one or more images from the training set.

Images with a low violation index or \textbf{min violating samples} are relatively unseen in the training data, which means that they may be novel or unique compared to other images. These images may contain features or characteristics that are not commonly found in the training set. On the other hand, images with a high violation index or \textbf{max violating samples} are more likely to closely resemble one or more images from the training set. This suggests that these images may contain familiar features that are present in the training data.

When considering the selection strategy in Active Learning, a naive solution could be to select the samples with minimum violation indices,i.e., the sampling technique $X$ gives us the set \(\{x\in{D_{U}}:\mid{\emph{VISet}_{D_{U}}}\cap(-\infty, \emph{VI}(E_{x})\mid<K\}\) where $\emph{VISet}_{D_{U}}$ is the set of violation indices for the images in the unlabelled set and $\emph{VI}(E_{x})$ is the violation index for image $x$ (Eq \ref{eq:4}). Although with this approach the perturbation in the existing embedding space is minimized, it loses out on the opportunity to reduce uncertainty by learning through closely resembling images. As such, there exists a tradeoff between reducing existing uncertainty and learning novel instances. Hence, a better selection strategy would be an ensemble of minimum and maximum violating image samples, i.e, a better $X$ gives us the set
\begin{align*}
    \{x\in{D_{U}}:\mid{\emph{VISet}_{D_{U}}}\cap(-\infty, \emph{VI}(E_{x}))\mid<p\}\,\,\,\,\cup\\
    \{x\in{D_{U}}:\mid{\emph{VISet}_{D_{U}}}\cap(\emph{VI}(E_{x}),\infty)\mid>(K-p) \}
\end{align*}
where $p \in I$, is a hyper-parameter such that \(0 \leq p \leq N_{U}\).

Since the violation index does not inherently capture the diversity of images, a further enhancement of our selection strategy includes diversity sampling to maximize novelty in the selected subset. To do this, we adopt a $kmeans++$ based clustering of the image embeddings followed by violation index-based selection inside each cluster. We select $kmeans++$ due to its ability to converge faster \cite{ash2019deep}. The overall workflow has been depicted in \autoref{fig:mainfig}.

\section{Experiments and Results}\label{sec:experiments}


\keypoint{Datasets:}We use QMUL-Shoe-V2 \cite{pang2019generalising, riaz2018learning} and QMUL-Chair-V2 \cite{song2018learning} datasets that have been specifically designed for FG-SBIR. QMUL-Shoe-V2 contains a total of 6,730 sketches and 2,000 photos, of which we use 6,051 and 1,800 respectively for training and the rest for testing. For QMUL-Chair-V2, we split it as 1,275/725 sketches and 300/100 photos for training/testing respectively. For each photo, one of its possible sketches is selected randomly, and is considered as its label. Initially, we consider 300 photo-sketch pairs as the labelled set and the rest as unlabelled (i.e. absence of their corresponding sketches).

\keypoint{Implementation:}We implemented our framework in PyTorch \cite{paszke2017automatic} accelerated by an 11 GB Nvidia RTX 2080-Ti GPU. ImageNet \cite{russakovsky2015imagenet} pre-trained VGG-16 \cite{simonyan2015very} network (embedding dimension $D=256$) is used as the backbone network for both sketch and photo branches. In all experiments, we use Adam optimizer \cite{kingma2014adam} with learning rate of $1e-4$, batch size $16$ and train the base model with a triplet objective. In the active learning setup, we conduct 5 cycles of complete training, where at the end of each round, we employ our sampling technique to add $K$ samples to the labelled pool after the provision of its actual sketch.

In each active learning round, we obtain the embeddings for the photos and sketches from the labelled set and the photos from the unlabelled set. Following this, our sampling technique utilises these embeddings to select $K$ photos from the unlabelled set, whose corresponding sketches from the dataset are then used to label them. Once labelled, these photos and their sketches are added to the labelled set.

\keypoint{Evaluation Metrics:}Following the standard FG-SBIR setting \cite{bhunia2020sketch, bhunia2021more}, we quantify the performance of the sampling technique using acc.@$q$ metric ($q=1, 10$), i.e. the percentage of sketches having true-match photos appearing in the top-$q$ list, after each AL round.



\subsection{Baselines}
To the best of our knowledge, there has been no previous work on active learning for fine-grained sketch-based image retrievals. Thus, we compare ours with a few SoTA active learning techniques adapted suitably for fine-grained SBIR. We choose three widely used baseline sampling techniques, namely: random sampling, kmeans sampling, and coreset sampling. As evident from our choice of baseline methods, we do not consider uncertainty-based sampling techniques due to the limitations discussed in Section 3.3. Random sampling is a widely used baseline for evaluating active sampling techniques. Here, we randomly select photos for labelling based on our-predefined budget. In K-means sampling \cite{zhdanov2019diverse}, we cluster photos into as many clusters as the labeling budget and select the photos closest to the cluster centroid. In case of a tie, we randomly select any one of the closest photos. Finally, Coresets sampling \cite{sener2017active} is another diversity sampling technique in which we utilise "coresets" to find the most representative photos for sampling. To find the coresets, we use a farthest-first approach to select maximally distant photos in the embedding space.

    
    

\subsection{Performance Analysis}\label{sec:perf}

\begin{figure*}
    \begin{center}
        \scalebox{.5}{\input{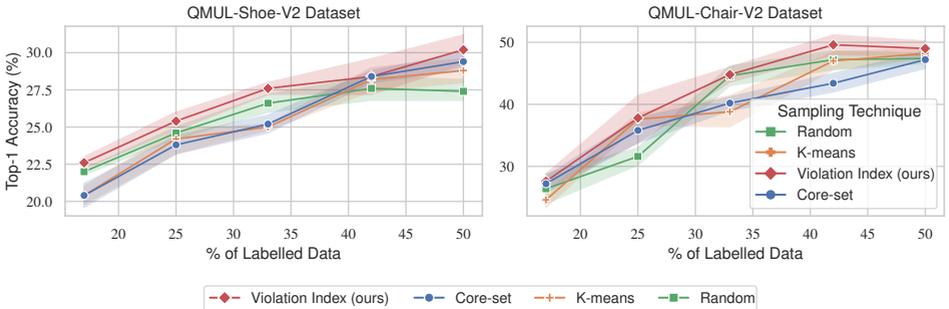}}
    \end{center}
    \vspace{-0.6cm}
    \caption{Comparing VI-based active sampling (ours) with SoTA AL baselines for fine-grained SBIR. All results are reported as mean of 5 runs and an error band of $\sigma=1.0$.}
    \label{fig:main}

\end{figure*}

To compare our approach with baselines, we report the mean and one standard deviation of acc.@$1$ values across 5 different runs. We also report the initial accuracy of the model. Figure \ref{fig:main} compares our violation index-based approach and baseline techniques. Before any active learning, the acc.@$1$ after training it on 8\% labelled data were 13\% (49.8\%) and 11.7\% (42.8\%) on the Shoe and Chair datasets respectively.

Our method outperforms the baselines on both QMUL-Shoe-V2 and QMUL-Chair-V2 datasets and obtains consistently higher mean accuracies. We consider the mean accuracy differences between our approach and the baselines to further substantiate the results shown in Figure \ref{fig:main}. On the QMUL-Chair-V2 dataset, we achieve a mean gain of 2.6\% acc.@$1$ compared to the 3 baseline techniques. On the QMUL-Shoe-V2 dataset, we observe a mean increase of 1.1\% acc.@$1$. 

Our proposed approach achieves comparable performance by utilizing only 40-50\% of the dataset. Compared to the state-of-the-art acc.@$1$ of 36.47\% on 100\% of the QMUL-Shoe-V2 dataset \cite{pang2019generalising}, we obtain a mean acc.@$1$ of 29.8\% by only utilizing 40\% of the dataset. This is achieved when we use $\alpha$=0 as our hyperparameter. We also see consistent results on the QMUL-Chair-V2 dataset (using $\alpha$=0.7), where we obtain 49.6\% acc.@$1$ by only using 50\% of the dataset. 



\subsection{Ablation Study}\label{abla}
To better understand the effects of violation index-based sampling, we perform ablations on violation index-based selection and the choice of hyper-parameter.

\keypoint{Significance of violation index and diversity-based sampling:}For this, we consider selecting only the unlabelled images that have the smallest and largest violation indices in active learning rounds performed on the QMUL-Shoe-V2 dataset.

\begin{wrapfigure}{r}{1\textwidth}
\centering
\includegraphics[width=1\textwidth]{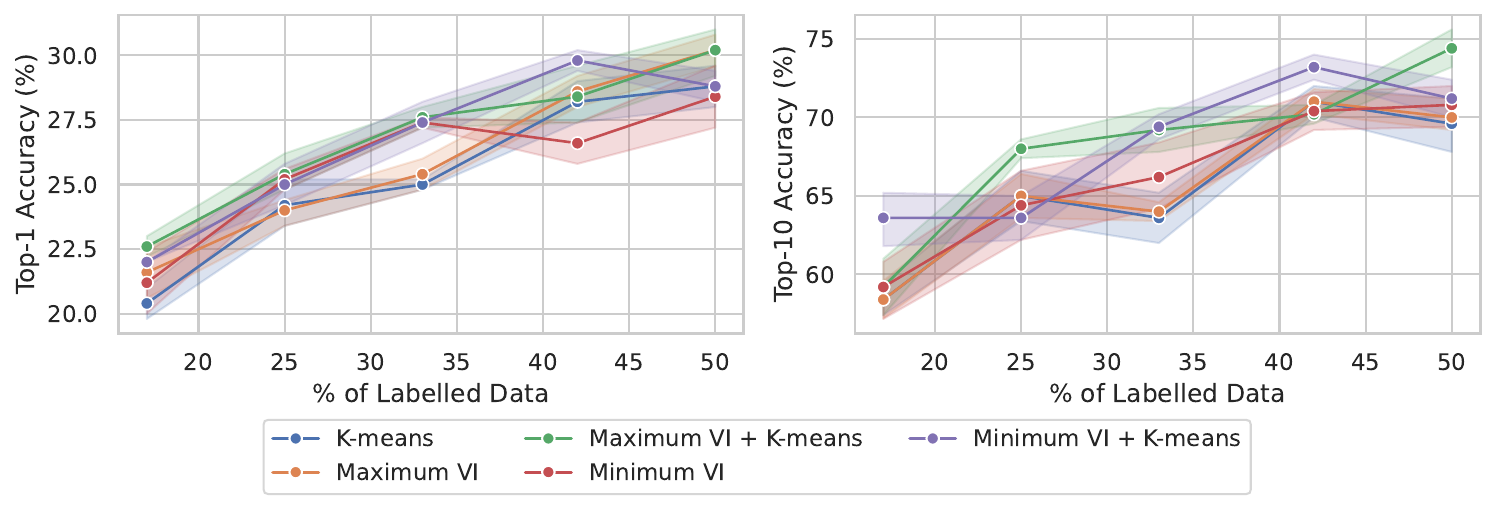}
\vspace{-0.5cm}
\caption{Ablations on diversity-based sampling and violation index. We compare only VI-based sampling, diversity-based sampling, and a combination of both.}
\label{fig:div}
\end{wrapfigure}

As per Figure \ref{fig:div}, an interesting observation is made: In early active learning cycles, selecting minimum VI yields higher acc.@$1$ compared to maximum VI. As the training dataset grows, this relationship is inverted, with higher acc.@$1$ achieved through Maximum VI-based selection. Empirical results support our hypothesis on violation indices, indicating that with less training data, increased retrieval accuracy primarily stems from selecting diverse instances. Conversely, with larger training data, reducing model uncertainty on similar instances contributes more to acc.@$1$. Thus, selecting maximum VI samples yields optimal results.

\vspace{-1.5cm}

\begin{wrapfigure}{r}{0.6\textwidth}
\centering
\includegraphics[width=0.62\textwidth]{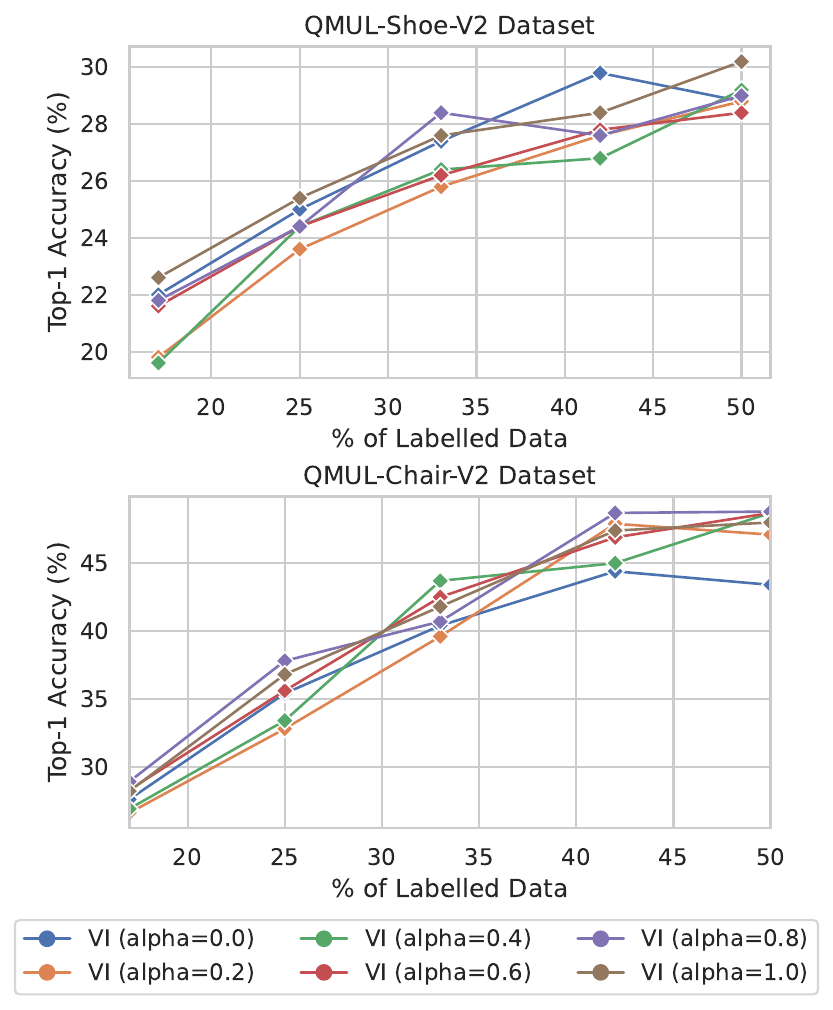}
\vspace{-0.6cm}
\caption{Ablations on value of hyper-parameter $\alpha$ on the ShoeV2 and ChairV2 datasets.}
\vspace{-0.9cm}
\label{fig:vi}
\end{wrapfigure}

Another significant aspect of our experiment involves the need for diversity-based sampling, not just violation index-based sampling. As shown in Figure \ref{fig:div}, violation index-based selection within diverse clusters obtained from kmeans++ consistently outperforms vanilla violation-based approach and kmeans++. Thus, the violation index captures semantic relations between labeled and unlabeled datasets but does not explicitly utilize relations between unlabeled images for diverse selection.

\begin{figure}[H]
\centering
\includegraphics[width=1\columnwidth]{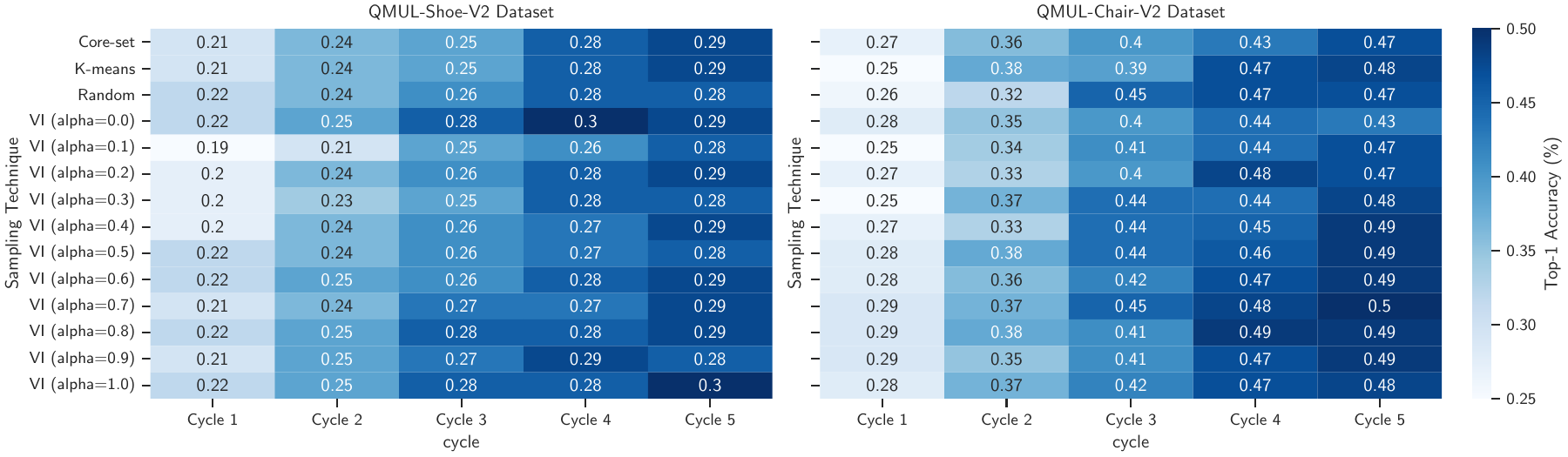}
\vspace{-0.5cm}
\caption{Ablations on value of hyper-parameter $\alpha$ and its comparison with baselines on the QMUL-Shoe-V2 and QMUL-Chair-V2 datasets. }
\label{fig:heat}
\vspace{-0.2cm}
\end{figure}

\keypoint{Sensitivity to hyper-parameter: } We analyze the sensitivity of our method to the hyperparameter $\alpha$, the results shown in \autoref{fig:vi}. We vary the value of  $\alpha$ from 0 to 1 with a gap of 0.1 and report the mean acc.@$1$ on both datasets. On the QMUL-Shoe-V2 dataset, we observe a steady increase in acc.@$1$ as we increase $\alpha$. On this dataset, $\alpha$=0 produces the best results. On the QMUL-Chair-V2 dataset, we observe a steady increase in initial and final acc.@$1$ as we increase $\alpha$. As per Figure \ref{fig:heat} and on this dataset, $\alpha$=0.7 produces the best results. On both datasets, there is a sudden drop in performance as we increase $\alpha$ from 0.0 to 0.1 and 0.2. We also present the same results compared with the baseline method in Figure \ref{fig:heat}. We believe that studying this interesting behavior is an open research avenue.

\section{Conclusion}
We have proposed an active learning framework to tackle the annotation bottleneck in fine-grained SBIR. To this end, we brought forth a quantifiable metric, \textit{violation index}, that measures the latent space displacements due to addition of a new instance. With suitable experiments and ablations, we have shown the robustness of our model compared to classification-specific AL works, especially in the low-data regimes. Our model is modality agnostic and thus can be leveraged for any cross-modal retrieval task. In future, we plan to extend our studies by investigating its applicability to such other tasks.

{\small
\bibliography{Original_egbib}
}

\end{document}